\documentclass[a4paper,journal,10pt, onecolumn]{IEEEtran} 
\usepackage[utf8x]{inputenc}

\usepackage{setspace} 
\usepackage{cite}
\ifCLASSINFOpdf
   \usepackage[pdftex]{graphicx}
  \graphicspath{{img/}}
  \DeclareGraphicsExtensions{.pdf,.jpeg,.png}
\else
\fi

\usepackage[cmex10]{amsmath}

\usepackage{algorithmic}

\usepackage{array}

\usepackage{fixltx2e}

\usepackage{stfloats}
\fnbelowfloat

\usepackage{url}

\begin{document}
%
\title{Fuzzy Sets Across the Natural Language Generation Pipeline}

\author{\IEEEauthorblockN{A. Ramos-Soto}, \IEEEauthorblockN{A. Bugarín}, \IEEEauthorblockN{S. Barro}\\
\IEEEauthorblockA{Centro Singular de Investigaci\'on en Tecnolox\'ias da Informaci\'on (CiTIUS)\\
Universidade de Santiago de Compostela\\
Santiago de Compostela, Spain\\
\{alejandro.ramos,alberto.bugarin.diz,senen.barro\}@usc.es}}

\maketitle

\begin{abstract}
We explore the implications of using fuzzy techniques (mainly those commonly used in the linguistic description/summarization of data discipline) from a natural language generation perspective. For this, we provide an extensive discussion of some general convergence points and an exploration of the relationship between the different tasks involved in the standard NLG system pipeline architecture and the most common fuzzy approaches used in linguistic summarization/description of data, such as fuzzy quantified statements, evaluation criteria or aggregation operators. Each individual discussion is illustrated with a related use case. Recent work made in the context of cross-fertilization of both research fields is also referenced.\footnote{This paper encompasses general ideas that emerged as part of the PhD thesis ``Application of fuzzy sets in data-to-text systems''. It does not present a specific application or a formal approach, but rather discusses current high-level issues and potential usages of fuzzy sets (focused on linguistic summarization of data) in natural language generation.}

\end{abstract}


%
\IEEEpeerreviewmaketitle

\section{Introduction} \label{sec:intro}
\begin{figure}[h]
\centering
\includegraphics[width=0.8\columnwidth]{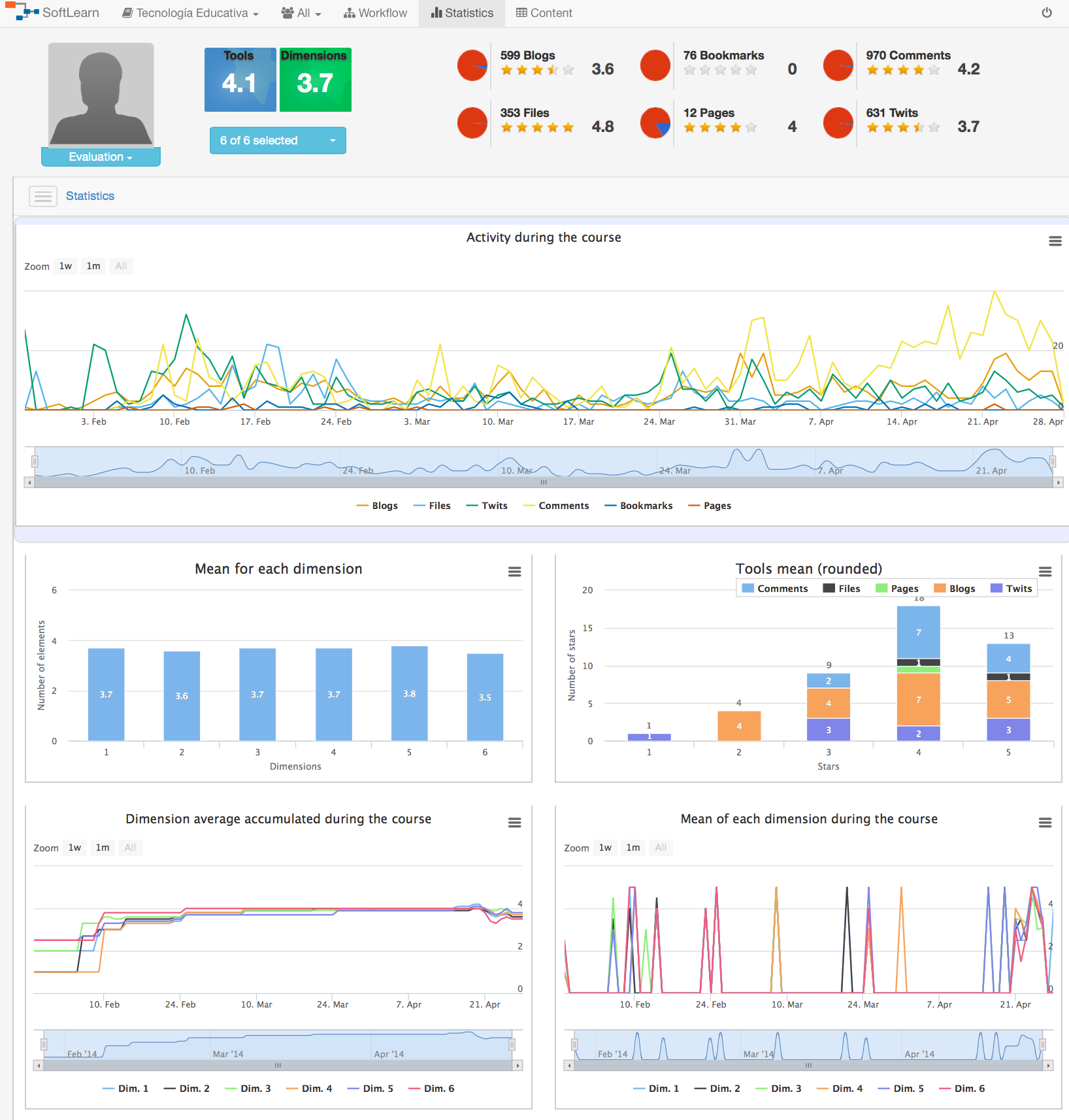}
\caption{Snapshot of the learning analytics dashboard SoftLearn \cite{softlearn_demo,softlearn_paper}, which displays metrics and data plots a teacher must interpret to assess how students perform in a course.}
\label{softlearn}
\end{figure}

Data Science has traditionally relied on analytics and visualization techniques to make sense of large volumes of data. Data scientists employ different techniques such as statistics, signal processing, pattern recognition, data mining or machine learning among others to extract relevant information from such amounts of data. However, communication of the extracted information after the analytics process is usually made through graphics or visualization techniques that in general demand interpretation efforts from the user side and sometimes require a rather extensive academic development or expertise for its actual comprehension (Fig. \ref{softlearn} shows an example of this problem).

This issue arises the interest of using other kind of complementary descriptive techniques which help fill the gap between data and users in a more human-friendly way, so that the obtained information can be grasped by a wider range of people regardless of their expertise. In this regard, approaches such as linguistic description of data (LDD) or natural language generation (NLG), which provide information expressed in terms of natural language, have emerged as feasible complements which, while still exploiting the full potential of standard Data Science analytics, allow for a better understanding of what underlies in such data. In this regard, recent studies \cite{nlg_metofficedatatotext} indicate that non-specialized users actually strongly demand textual descriptions of data as a means for better understanding of graphics and visualizations.

The discipline of linguistic description or summarization of data (LDD) \cite{bib_Yager} has become in recent times a very promising approach to capture the essential information residing in numerical data. It allows to easily obtain linguistic information structured in protoforms able to describe relationships between the values of the different input variables along time and even space dimensions. Linguistic description of data is based primarily on the use of fuzzy quantified sentences and its main contribution is that it allows to manage the uncertainty and vagueness present in the human language concepts which are used to summarize data. This has led to an apparition of a very diverse collection of use cases in very different domains (a rather representative review of such approaches can be found in \cite{bib_role_ldd_nlg,bib_ldd_time_series}).

At the same time, the natural language generation (NLG) field \cite{nlg_building_nlg}, which addresses the creation of systems able to automatically deliver texts indistinguishable from those produced by humans, is currently experiencing a bursting scientific, technical and commercial expansion in its data-to-text (D2T) specialty \cite{nlg_datatotext,bib_role_ldd_nlg} due to the rise of the Big Data era. The more data is available, the more time experts and users need to make sense of it and, while it may be a mundane task, the creation of reports that describe in a few paragraphs what in origin were huge amounts of data is usually necessary in any organization. In this regard, D2T solutions help analysts, experts and users in general in saving time by performing data analysis and deliver relevant information as high quality texts.

It is safe to assume that the D2T solutions provided by NLG companies do not include any uncertainty or vagueness management \cite{nlg_datatotext}. In fact, although NLG (and D2T by extension) excels in terms of generating texts whose quality is optimal from a linguistic perspective, the problem of how to address vagueness is still an open issue which is being actively researched in this discipline \cite{nlg_vandeemter06,nlg_vandeemter09,nlg_power12,nlg_impreciseportet}.

In this context, the current state of both LDD and NLG fields has led to a climate of mutual interest. LDD approaches may use NLG techniques to convert linguistic protoforms into information in an even more human-friendly state, which allows the delivery of high quality texts. Likewise, NLG systems may use LDD and other fuzzy-related techniques to address the problem of uncertainty at different levels within the distinct tasks involved in an NLG process. Examples from both trends include the GALiWeather \cite{bib_galiweather} system, conceived as a LDD approach but made use of several NLG techniques to generate short-term weather forecasts; and the model presented in \cite{bib_gatt_portet_uncertain} that involves fuzzy temporal constraint networks and experimental data from three languages to address the problem of generating uncertain temporal expressions from an NLG point of view.

The previous examples show that collaboration between both fields is slowly but steadily starting to take place. However, it is yet unclear which and how many forms this potential cross-fertilization process will take. In this situation, the aim of this paper is to study from a detailed and practical perspective several potential convergence points for LDD and NLG which may open new research lines in the context of such collaboration. Furthermore, we will also resume the discussion opened by Kacprzyk and Zadro\.zny in \cite{bib_kacprzyk_4,kacprzyk_n2}, where a relationship between LDD and NLG was discussed for the first time at a conceptual level, and which was later expanded by \cite{bib_role_ldd_nlg,bib_ldd_time_series}, which provided additional insights and state of the art reviews for both fields.

This paper is organized in three main sections. Section \ref{sec:lddandnlg} explores the challenges that LDD currently faces and how its usage together with NLG can help in addressing them. Section \ref{sec:nlgpipeline} studies in depth how the use of LDD techniques (namely fuzzy quantified statements) relate to the NLG tasks described by the well-known and widely accepted NLG system architecture proposed by Reiter and Dale in \cite{nlg_building_nlg}, as well as potential implications derived from such usage. Finally, Section \ref{sec:conclusions} provides further insights into some of the topics of most interest in the context of the collaboration between LDD and NLG and summarizes the main contributions of this paper.

\section{LDD and NLG: Main Challenges} \label{sec:lddandnlg}
Linguistic descriptions of data can be used to extract relevant information from numeric input data sets. Such descriptions are usually based on the concept of fuzzy quantified statement \cite{bib_Yager}, which is usually classified using Zadeh's notion of protoform \cite{bib_zadeh_5}. In this regard, two basic protoforms are distinguished
\begin{equation}
``Q \  Xs \  \text{are} \  A"
\label{type1}
\end{equation}
\begin{equation}
``Q \  DXs \  \text{are} \  A"
\label{type2}
\end{equation}
where $Q$ is a fuzzy quantifier, $A$ is a summarizer and $D$ is a qualifier. These protoforms are also a representation of fuzzy quantified statements commonly referred to as type-1 (Eq. \ref{type1}, ``a few researchers are young'' or ``some of the humidity values were high'') and type-2 (Eq. \ref{type2}, ``most of the cold days were very humid'').

Although protoforms have also been used in the context of fuzzy queries \cite{bib_kacprzyk_4,kacprzyk_n2}, we will focus on their application in the automatic extraction of relevant linguistic information, that is, approaches applied to practical use cases in particular knowledge domains, where the structure of the linguistic descriptions which are obtained is known a priori. Following this definition, fuzzy quantified statements in the literature have been used to a great extent as a means to summarize and describe time series of data \cite{bib_ldd_time_series}.

\begin{figure}[h]
\centering
\includegraphics[width=0.6\columnwidth]{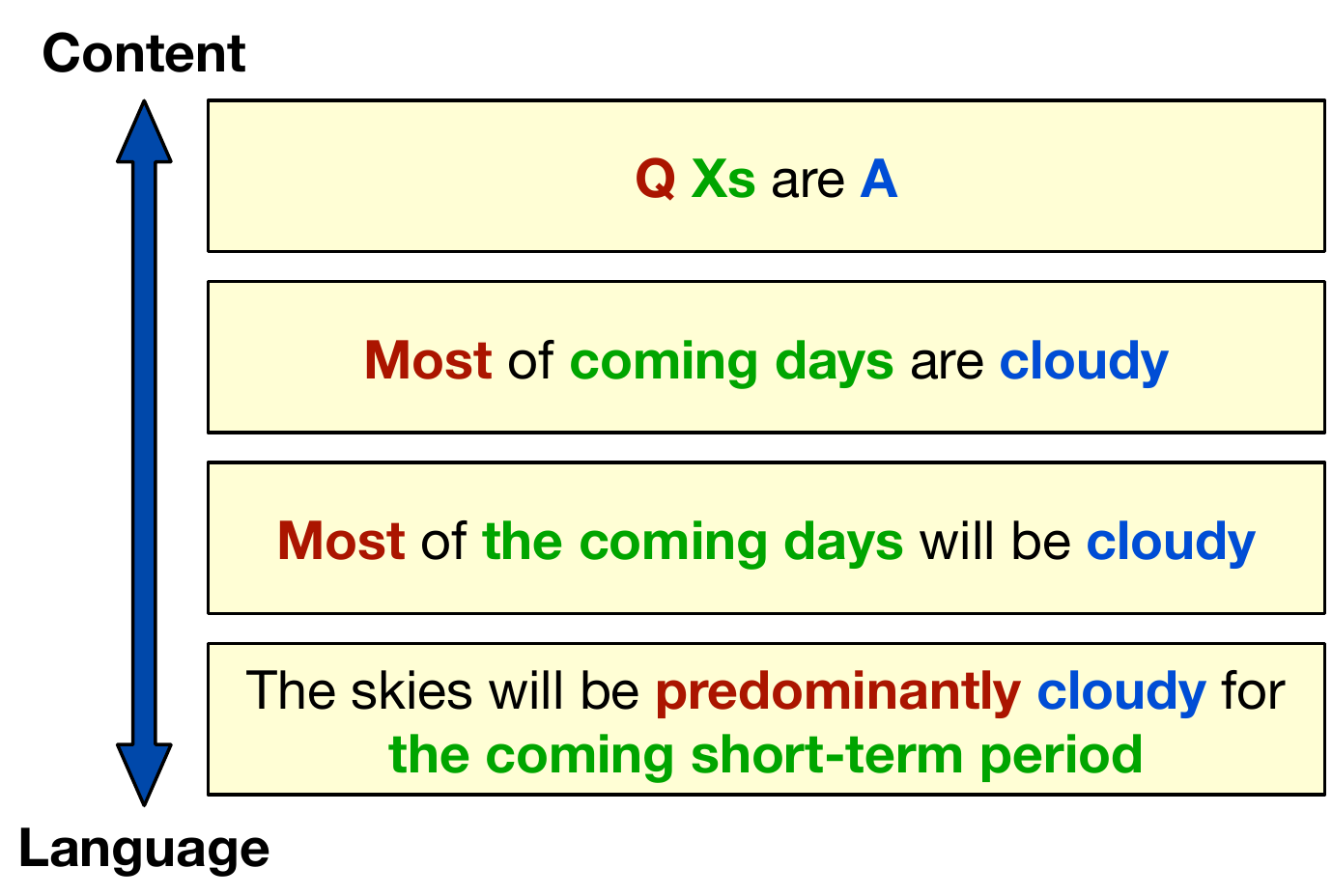}
\caption{Contrast between protoform-like linguistic information and an actual natural language text ready for human comsumption by general public.}
\label{protovstext}
\end{figure}

In this sense, LDD emerges as a discipline with a similar purpose to that of D2T approaches: to provide an understandable interface between the data and the human users in the form of information expressed in terms of natural language. However, data-to-text is aimed at the production of actual texts, while linguistic description of data remains in a more conceptual level. Figure \ref{protovstext} illustrates this contrast between both fields: while it is feasible to obtain a linguistic description which summarizes a data set properly, it is arguably useful for a human user if such linguistic information is not given in a way that matches the language used in the user's specific application domain.

Thus, the main downside of linguistic descriptions of data is that they are restricted to protoform structures and, although they are flexible enough to be extended into more sophisticated forms (including time and spatial dimensions, for instance), their usage in real applications is not feasible when provided as is. This issue was identified by Kacprzyk and Zadro\.zny in \cite{kacprzyk_n2}, where they stated that \textit{``Zadeh’s protoforms are very powerful and convenient in CWW but may be a limiting factor in many real-world applications as their structure is too restricted, notably as compared to the richness of natural language''}.

At the same time, this necessity for a means to effectively use the fuzzy techniques developed in LDD in practical application domains has also opened up the possibility of resorting to NLG. In this regard, Kacprzyk and Zadro\.zny proposed in \cite{kacprzyk_n2} to define new types of protoforms \textit{``to make a full use of the power of NLG tools''}. In their opinion, to be able to generate statements that provide distinct and richer expressiveness than standard protoforms would expand the potential usages of LDD and, at a general level, NLG techniques could be used to merge the best candidate statements obtained by a search algorithm into a proper text.

In our opinion, expanding the current collection of protoforms would entail a significant advance for LDD in terms of expressiveness, which in turn could probably bring LDD a step closer to NLG, but not necessarily closer to its usage in real systems. In this regard, the idea of NLG systems producing texts solely based on fuzzy protoforms, however complex these may be, seems unrealistic for the needs of actual systems.

For instance, the textual weather forecast generator GALiWeather \cite{bib_galiweather} employs type-1 fuzzy quantified sentences to perform a global description of the cloud coverage variable, but it also uses different crisp approaches to extract the relevant information from other variables such as precipitation or temperature. The usage of such distinct techniques responds to the needs of the domain experts, who provided both the linguistic specifications and most of the domain knowledge required to build the system. Thus, LDD provides powerful tools to manage uncertainty and imprecision in the generation of linguistic expressions, but first it should be determined when their usage is appropriate. This issue is directly related to a relevant concept also noted in \cite{kacprzyk_n2}, namely the domain-modeling.

\begin{table}[h]
\centering
\caption{Main challenges in LDD and proposed solutions through its integration into NLG.}
\begin{tabular}{| p{0.3\textwidth} | p{0.6\textwidth} |}
\hline
Challenge & Solution  \\ \hline
\noalign{\smallskip} \hline
When to use fuzzy/LDD techniques as part of an NLG system? & Corpus analysis \cite{bib_mousam}: Determine which expressions in the language requirements can be modeled with protoforms. Determine if there is imprecision or uncertainty in the provided domain knowledge. \\ \hline
How to build fuzzy sets based on empirical knowledge? & Distinct approaches:
\begin{itemize}
\item Analysis of parallel corpora of texts and corresponding data \cite{bib_mousam}. Alternatively, automatically learn such definitions (for instance, as it is done in TSK rule-based systems \cite{bib_tsk}).
\item Perform experiments to obtain empirical definitions from experts or human users \cite{bib_rodrigo,bib_gatt_portet_uncertain}.
\end{itemize}
\\ \hline
\end{tabular}
\label{nlgtoldd}
\end{table}

Thus, the perspective should shift from ``how to use standalone LDD in real applications'' to ``when and how to use LDD as part of real NLG (or D2T) systems''. In a general sense, the analysis of the texts of a specific application domain (or the linguistic requirements of the experts if no example texts are available) will shed light on this issue\footnote{This technique is commonly known in NLG as ``corpus'' analysis and is performed systematically in the early conception stages of an NLG system \cite{bib_mousam}.}. If certain expressions in the language requirements match (directly or indirectly, from a content perspective) any of the protoforms currently defined and the linguistic terms to be used in such expressions allow for a fuzzy numerical definition (the domain experts cannot provide specific crisp definitions of such terms or there is not a consensus definition), then LDD could be properly applied.

This directly leads to another challenge that restricts the usage of LDD in real applications and which has not been previously considered, namely the problem of building fuzzy definitions of linguistic terms based on expert knowledge. In general, the problem of mapping the intuitive notion of a subjective concept from the application domain into a fuzzy set or relationship has not been a primary concern in the literature in LDD, as theory and use cases had to be developed first in order to show the potential applications of this kind of techniques. Even in more recent applied approaches which generate actual texts, linguistic variables were defined by authors and the quality of each solution as a whole was checked through an evaluation process by experts, e.g. \cite{bib_quality,bib_galiweather}.

While to impersonate an expert domain in order to fill knowledge gaps for the application of LDD techniques can be considered admissible and plausible to a certain extent, this practice seems to be in conflict with the purpose of a domain-modeling process: in order to be able to use LDD, the author creates fuzzy definitions for linguistic terms based on self-judgments about the application domain, rather than capturing this meaning from the domain itself. In this sense, NLG has traditionally used empirical techniques to assess the meaning of words and terms in an as accurate as possible way. For instance, for the development of the NLG system SumTime-Mousan \cite{bib_mousam}, Reiter et al. analyzed a parallel set of textual wind forecasts by five different experts and their corresponding data in order to achieve a coherent definition of temporal expressions such as ``by evening'' or ``by midday''. Subsequent evaluations of the system showed that overall forecast readers preferred the wind texts generated by the system over human-written wind texts. In other cases experiments were run in order to study how human subjects use linguistic expressions in different domains (e.g. \cite{bib_rodrigo, bib_gatt_portet_uncertain}).

In this regard, the main challenge resides in bringing such empirical approaches into LDD and adapting them for achieving a proper definition of fuzzy linguistic terms. This, depending on the kind of LDD statements which could be used, also opens up the possibility of performing experimentation to determine, for instance, which operators could be used effectively for combining different summarizers or properties (e.g. as in ``most of the students are short and fast''), such as the compensatory operators proposed by Zimmermman \cite{bib_zimmermann} or the OWA operators by Yager \cite{bib_yager_owa}. In a general sense, this would imply the instantiation of the theoretical models developed in fuzzy sets based on standard NLG empirical approaches for different application domains.

Although LDD faces important challenges (Table \ref{nlgtoldd} summarizes the main aforementioned issues and hints for addressing them), there is an important consensus in this field about its practical viability based on the use of NLG tools \cite{kacprzyk_n2,bib_role_ldd_nlg,bib_ldd_time_series,editorialfss}. This trend has been reinforced in recent years, as more work in LDD has adopted the use of template-based NLG approaches \cite{nlg_nlgvstemplate} to generate actual texts (e.g. \cite{bib_galiweather, bib_quality}).

The remaining question, now it is clear how LDD should be approached effectively in conjunction with NLG, resides in clarifying what benefits and implications LDD (and fuzzy sets in general) may bring to NLG systems. This was also noted by Kacprzyk and Zadro\.zny in \cite{kacprzyk_n2}: \textit{``...it seems that NLG can benefit from the approach and solutions that are adopted in our approach by finding the conceptually and numerically operational means to grasp and handle the problem of imprecision of meaning that is so characteristic for natural language but has not been appropriately considered in NLG despite an urgent need''}.

Imprecision and uncertainty in NLG have not been fully addressed, but it is a subject of interest and research, as stated in Section \ref{sec:intro} (e.g.
\cite{nlg_vandeemter06,nlg_vandeemter09,nlg_power12,nlg_impreciseportet}). However,
it is not clear why fuzzy sets and derived disciplines such as LDD have not been considered by the NLG community until very recently \cite{bib_gatt_portet_uncertain}, despite being intuitively appropriate for this task. This unawaraness may be partially explained by the opposition between the traditional theoretical nature of the fuzzy field, more focused on its logical aspects, and the more applied nature of NLG, focused on linguistic problems of a more empirical weight.

Nowadays, thanks to the effort and work made from both fields, we have a better understanding of where LDD fits in an NLG process and what benefits this relationship may bring in the task of textually describing large data sets while  also taking into account the management of uncertainty and imprecision. In this regard, Section \ref{sec:nlgpipeline} deals with this question in a thorough way from an NLG perspective.

\section{Fuzzy Sets and LDD Across the NLG Pipeline} \label{sec:nlgpipeline}

\subsection{The NLG Pipeline}
NLG systems can be generally depicted as systems tasked with the conversion of some input data into an output text. The most widely accepted classification of this task division is the architecture proposed by Reiter and Dale in \cite{nlg_building_nlg}, where NLG systems are characterized as a pipeline composed of several parts which deal with different aspects of the NLG process. In this pipeline NLG systems are informally structured as a process divided into three principal modules: \textit{i)} text or document planning, \textit{ii)} microplanning and \textit{iii)} realization. Document planning is focused in the production of a specification of the text's content and structure, microplanning addresses the problem of choosing appropriate expressions for the content and other fine-grained tasks and realization produces the actual text by applying grammatical, syntactical and ortographical rules.

\begin{figure}[h]
\centering
\includegraphics[width=0.7\columnwidth]{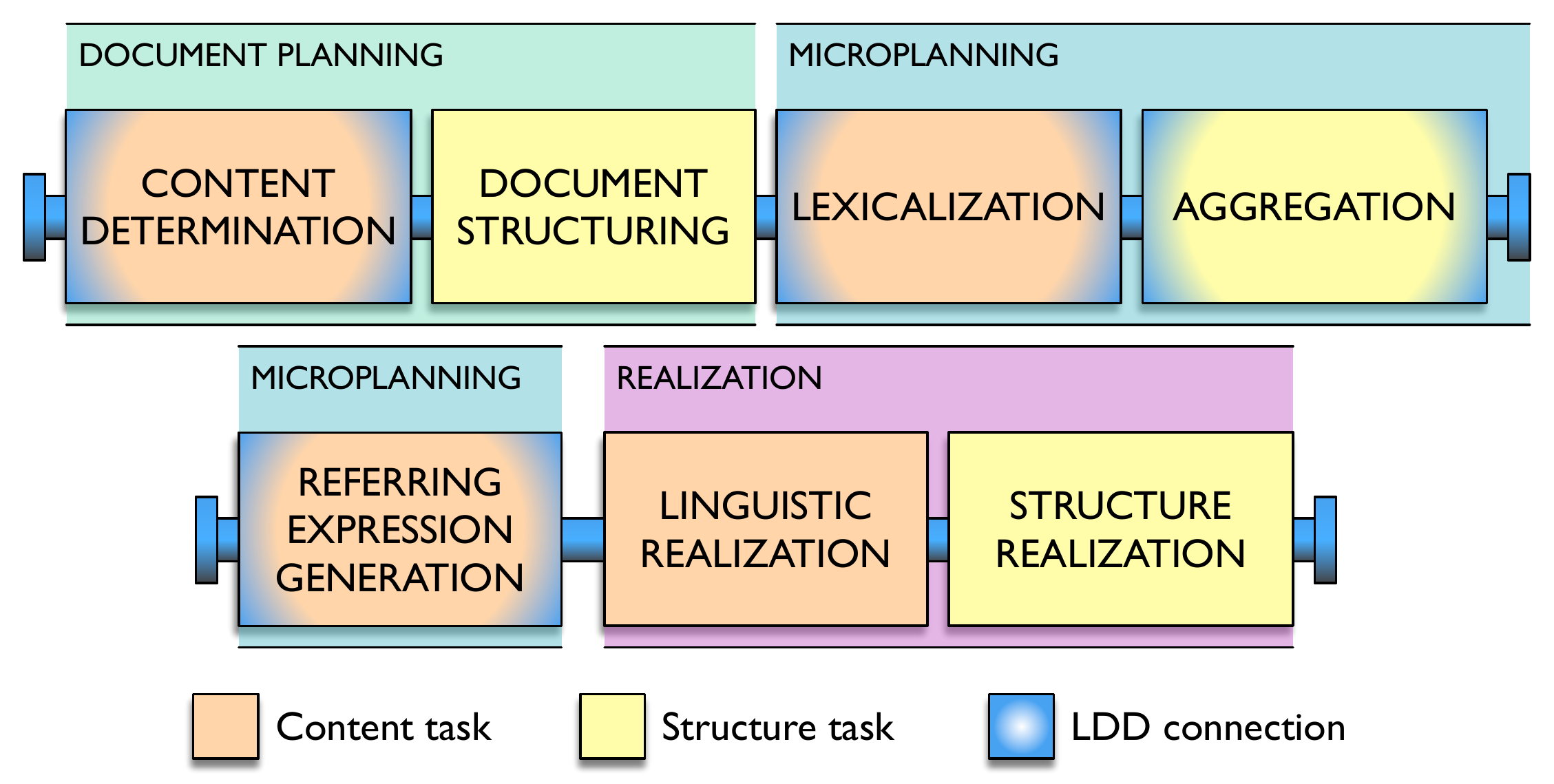}
\caption{Natural language generation task pipeline according to Reiter and Dale \cite{nlg_building_nlg}, and potential LDD connections.}
\label{pipeline}
\end{figure}

A more detailed division of such architecture distinguishes six different activities (see Fig. \ref{pipeline}). These subtasks focus alternatively on two different dimensions of the NLG process, namely content and structure:

\begin{itemize}
\item \textbf{Content determination}. As the name indicates, this process identifies and isolates the relevant information which will be communicated in the text from the source input data in the form of input messages.
\item \textbf{Document structuring}. Organizes the set of messages according to a certain order and structure, including how to group messages together thematically, their order or their correspondence to high level document structures such as paragraphs and sections.
\item \textbf{Lexicalization}. This process involves deciding which specific words and expressions shall be used to express the information obtained in the content determination task.
\item \textbf{Aggregation}. Similar to document structuring, this task is concerned with deciding how the information is mapped into low level structures such as sentences.
\item \textbf{Referring expression generation}. This task deals with the identification of entities in a discourse. This implies selecting the words or expressions that allow to identify such entities.
\item \textbf{Linguistic realization}. Converts sentence representations into actual text. 
\item \textbf{Structure realization}. Converts the high level structures into mark-up symbols understood by the document presentation component (for instance, if paragraphs or sections are part of an HTML file which is used to display the output text).
\end{itemize}

Of all the subtasks described in \cite{nlg_building_nlg} by Reiter and Dale, content determination emerges as the more intuitively related to LDD. In this sense, the current consensus in LDD is that this kind of techniques allow to perform a ``fuzzy'' content determination task. Kacprzyk and Zadro\.zny in \cite{kacprzyk_n2} introduced this notion in terms of the more general concept of text planning, while \cite{bib_role_ldd_nlg,bib_ldd_time_series} explored in more detail the role of LDD as a means to perform content determination while managing at the same time the problem of imprecision in natural language.

This has opened up new ways for further exploration of the relationship between both fields, which allow to consider an even wider usage of fuzzy sets and LDD beyond content determination. In this regard, we are also considering lexicalization, aggregation and referring expression generation as subtasks which can be addressed or influenced by means of LDD (Fig. \ref{pipeline}). We will illustrate this through a use case inspired by both the \textsc{WeatherReporter} design study provided in \cite{nlg_building_nlg} and the LDD example presented in \cite{fuzzieee_model}, where a generic model to approach LDD in the context of performing content determination tasks was proposed.

\subsection{An Illustrative Use Case}

Let us suppose there is a meteorology agency which provides weather reports based on observational data series obtained from several meteorological sensor stations spread across 100 different locations in a specific region. This agency is interested in automatizing the creation of reports, as this task involves a lot of analysis time and effort from the meteorologists due to the high number of reports which must be produced (one per location).

\begin{figure}[h]
\centering
\includegraphics[width=0.8\columnwidth]{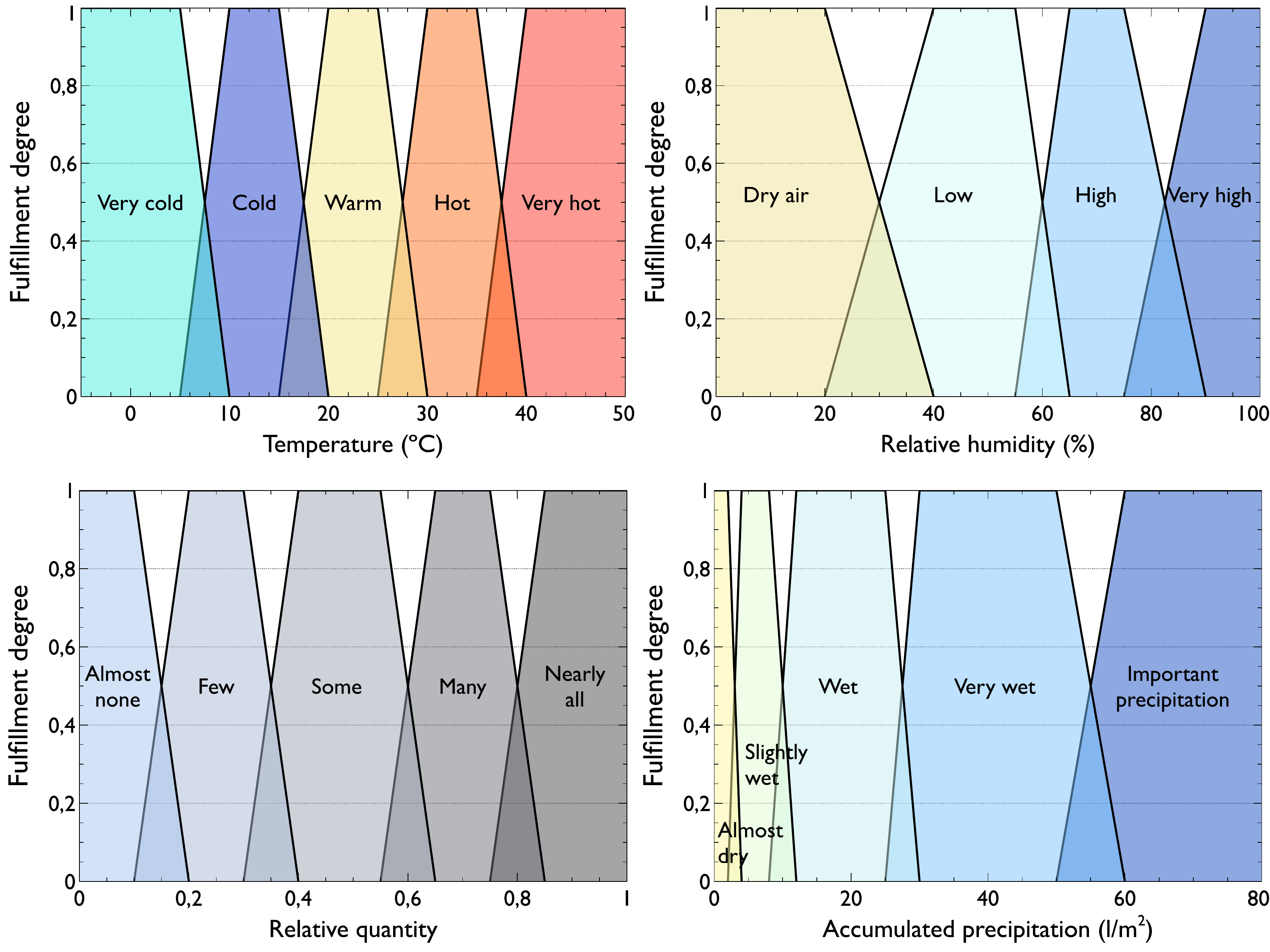}
\caption{A fuzzy knowledge base for the use case in Section \ref{sec:nlgpipeline}, including linguistic variables and a partition of fuzzy quantifiers.}
\label{knowledge}
\end{figure}

A team composed of NLG experts is then tasked with the development of a system which generates meteorological textual reports. After a few preliminary meetings and studying some examples of parallel human-produced texts and corresponding data, the experts determine that the specifications provided by the meteorologists are incomplete and imprecise: in some cases there is not a clear semantic definition of the linguistic terms to be used in the text generation process and meteorologists seem to diverge in their definitions.

A few experts in fuzzy sets and their application in LDD join the NLG system development team to help them address and model the imprecision found during the initial stages of the domain-modeling process. The team then designs several experiments, which are performed by the meteorologists, focused on the semantics of the terms related to the meteorological variables of interest, namely the temperature, humidity and precipitation, and other terms such as quantifiers. Based on the obtained empirical data, the team is able to build a fuzzy knowledge base such as the one shown in Fig. \ref{knowledge}. This allows to start the design and development of the NLG system, which will include expressions produced through the use of fuzzy and LDD techniques.

\begin{figure}[h]
\centering
\includegraphics[width=0.7\columnwidth]{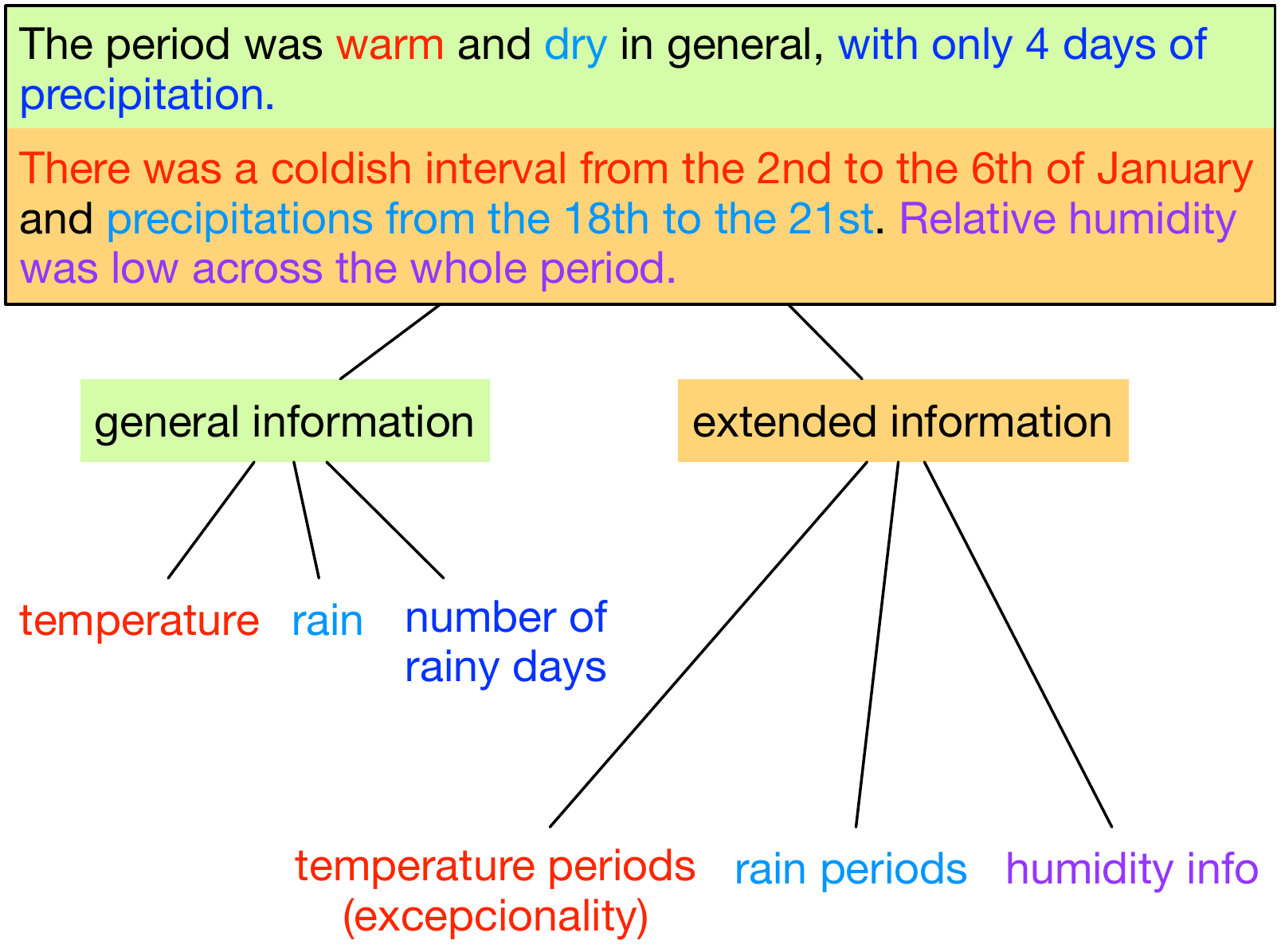}
\caption{Example of a report which should be generated by the NLG system and its structure.}
\label{example}
\end{figure}

In a general sense, the NLG system will receive as input a data set composed of time series data for several variables from a given meteorological station in a given time period and generate a textual report as output. The reports generated by the system will include both general facts and more specific details about the behavior of the variables of interest (see Fig. \ref{example}). In this context, we will study how fuzzy sets and LDD may interact with the NLG tasks.

\begin{figure}[h]
\centering
\includegraphics[width=0.8\columnwidth]{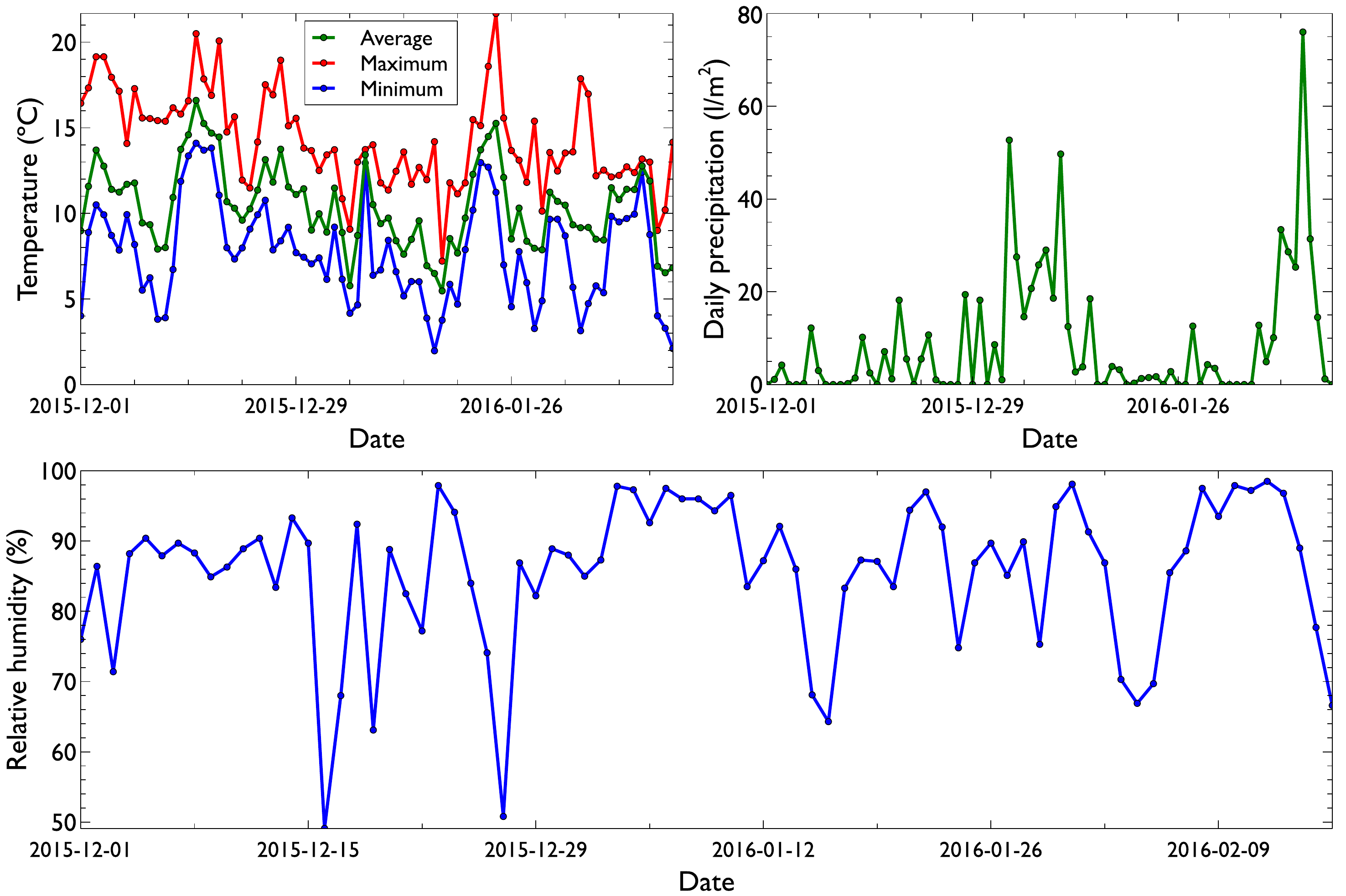}
\caption{Meteorological observation data for this use case.}
\label{weather_data}
\end{figure}

\subsection{Across the NLG Pipeline} \label{sec:across}

\subsubsection{Content determination} \label{sec:cd}
This task is tied to the specific application domain, as it involves employing different techniques to extract relevant information from the source data. The problem of which techniques to use depends on the nature of the source data and the content requirements of the texts to be produced. For instance, in the present use case the NLG system development team could apply signal processing or pattern recognition techniques, as the system will have to deal with time series data for several variables (see Fig. \ref{weather_data}). Although the content requirements in this use case are more restricted for illustration purposes, actual NLG systems do include a wide variety of this kind of approaches. This is especially true in the case of D2T systems, such as BabyTalk \cite{nlg_neonatal,nlg_babytalk2} or SumTime-Turbine \cite{nlg_turbine,nlg_turbine2}).

\begin{figure}[h]
\centering
\includegraphics[width=0.4\columnwidth]{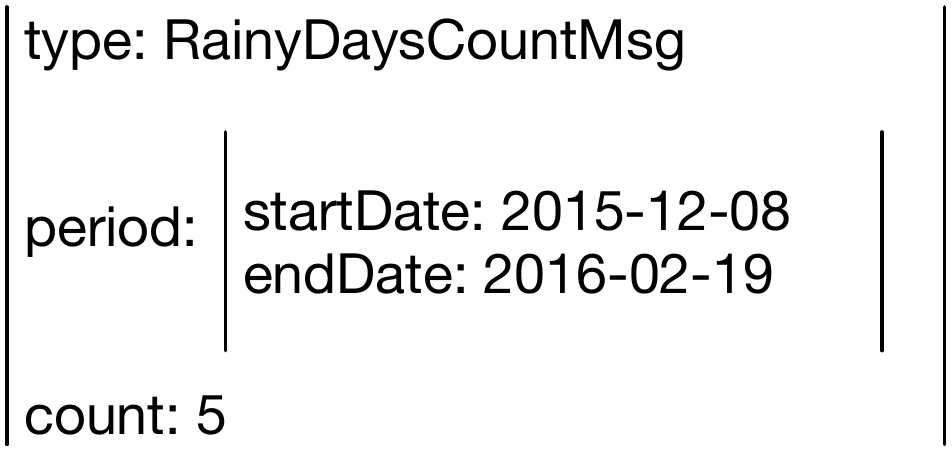}
\caption{Example of a content determination message for the rainy days count element.}
\label{rain_count_message}
\end{figure}

If we focus on the content which must be included in the ``general information'' part of the meteorological report, three main elements can be identified (Fig. \ref{example}). The first and second ones provide information about the predominant trend for temperature and rain in a given time period, while the third one is just a count of the number of days where precipitation was registered. Following the concept of message described in \cite{nlg_building_nlg}, Fig. \ref{rain_count_message} shows a possible content determination message for the rainy days count.

Let us suppose then that the first two elements which describe the temperature and rain trends in the human-produced reports are a vague description based on a subjective quantification performed by the meteorologists. The fuzzy experts determine that, based on the fuzzy definitions shown in Fig. \ref{knowledge}, these elements can be modeled using type-I fuzzy quantified statements, such as ``most of the days the temperature was cold'' or ``a few days were wet''. This increases both the flexibility and complexity of the content determination task for temperature and rain, which is developed by the fuzzy experts using the following strategy:
\begin{itemize}
\item All posible combinations of fuzzy quantifiers and labels are computed for each content element (e.g., from ``few, very cold'' to ``nearly all, very hot''). This involves determining the fulfillment degree (FD) of each quantified statement and additional criteria, such as the coverage degree (Cov) of the quantifiers \cite{bib_felixisda}.
\item The best candidate statements are selected based on the chosen criteria. In our use case, the fuzzy experts decide to choose the minimum number of candidate statements with a high fulfillment degree while covering as much data as possible at the same time.
\item More than one statement can be selected depending on the computed criteria. For instance, a ``nearly all, cold'' statement with a 0.9 fulfillment degree would suffice to describe the temperature, since ``nearly all'' is the quantifier with highest coverage and the fulfillment degree is high. However, both ``some, almost dry'' with a 0.6 fulfillment degree and ``some, wet'' with a 0.4 fulfillment degree would be selected as content messages, since the sum of both fulfillment degrees is the highest and the added coverage of both quantifiers is compatible with 100\% of the data.
\end{itemize}

\begin{figure}[h]
\centering
\includegraphics[width=0.85\columnwidth]{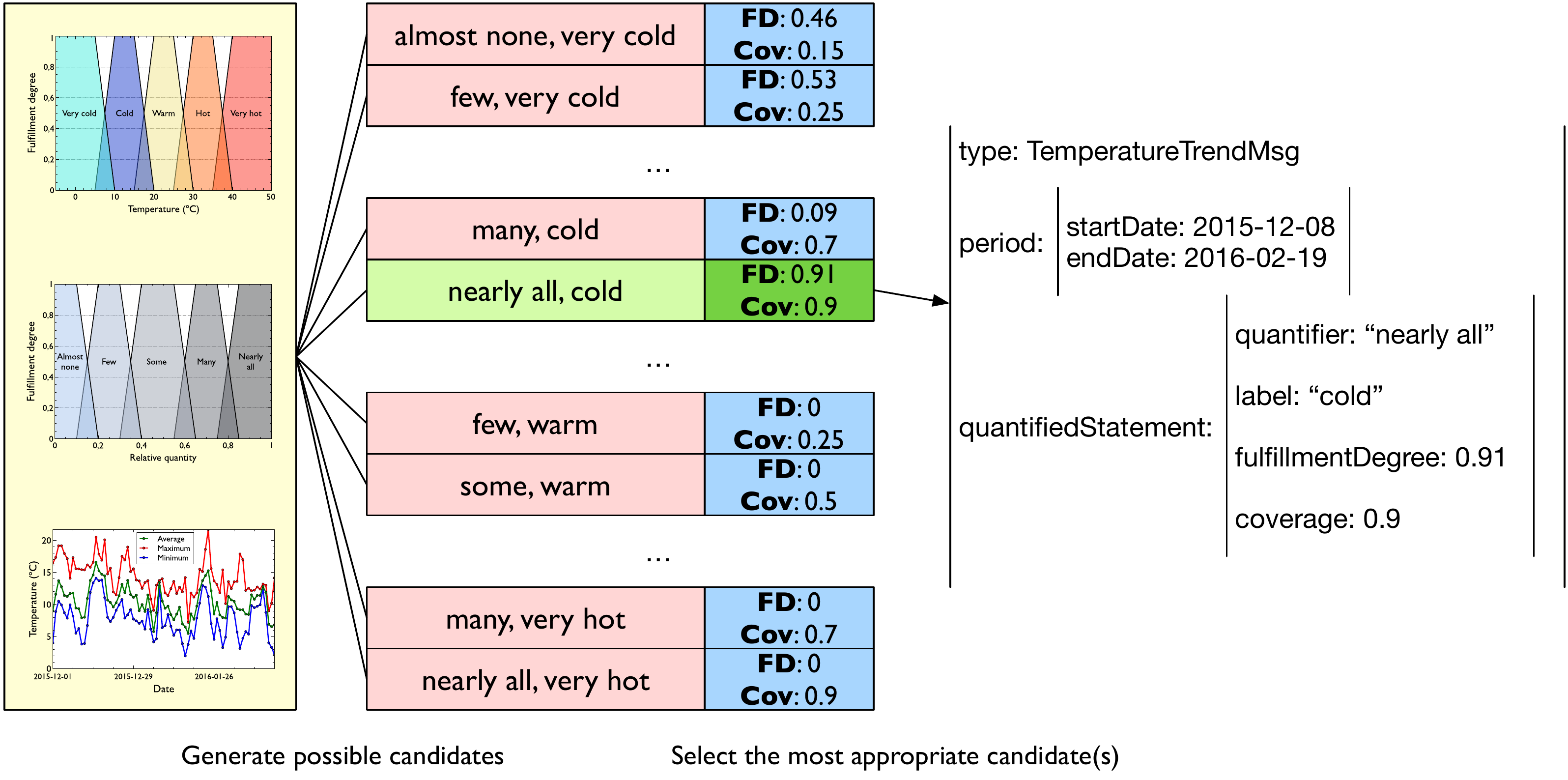}
\caption{Illustration of a content determination process based on type-I fuzzy quantified statements for the temperature element. Fulfillment degrees were calculated using Zadeh's approach \cite{bib_zadeh_quant} from the data shown in Fig. \ref{weather_data} according to the labels defined in Fig. \ref{knowledge}.}
\label{content_determ_quant}
\end{figure}

The chosen fuzzy quantified statements are then included as part of the content determination messages, such as the rainy days count message (Fig. \ref{rain_count_message}). This process is illustrated by Fig. \ref{content_determ_quant}, which also shows an example of a content determination message produced as a result of the fuzzy quantification algorithm. In general terms, such kind of messages should include at least the fulfillment degree associated to the obtained piece of content, as well as any other element which characterizes this content (e.g. linguistic quantifier and label in the case of type-I quantified sentences), as these will play an important role in subsequent NLG tasks (see Section \ref{sec:lex}).

There are also further additional fine-grain issues which must be considered. In this use case, for instance, to choose a fuzzy quantification model is another decision the fuzzy experts had to make in order to calculate the fulfillment degree of the fuzzy quantified sentences. Particularly, in the previous calculations, Zadeh's proposal was used \cite{bib_zadeh_quant}
\begin{equation}
FD(``Q \ Xs \ \text{are} \ A") = \mu_Q \left ( \frac{1}{n} \sum \limits_{i=1}^n \mu_A (x_i) \right )
\label{zquant_method}
\end{equation}
where $\mu_Q$ is the function associated to the fuzzy quantifier $Q$ (e.g. ``nearly all''), $\mu_A$ is a function associated to the  fuzzy label $A$ (e.g. ``cold''), and $X$ is the input time series data.

However, it is not clear which models might be the most appropriate for usage in real applications. Although we are aware of their theoretical properties \cite{bib_quantstate}, which can be useful in this regard, this remains an open problem which should be solved in each specific case \cite{bib_quantproperties}. This issue also applies to t-norm or t-conorm operators, which can be used to aggregate different properties (e.g. as in ``most of the days were cold \textbf{and} wet''). Minimum and maximum are perhaps the most commonly used, but there is a wide variety of operators which make such decisions harder to make (this will be further discussed in Section \ref{sec:agg}).

In general terms, using LDD or fuzzy approaches in content determination will imply creating a higher number of content messages which account for the imprecision defined in the fuzzy knowledge base: a discourse element such as the temperature or the rain will not be necessarily characterized by just one piece of information, but by several which contain fuzzy-related information (e.g. fulfillment degree, quantifier coverage, etc.) and which must be carried on to the following subtasks of the pipeline in order to be properly exploited. Particularly, lexicalization will be highly influenced by the decisions made in this initial task.

\subsubsection{Lexicalization} \label{sec:lex}

If content determination focuses on obtaining relevant information from the input data, the purpose of the lexicalization task is to move such content closer to an actual text. It is in this stage where the imprecision modeled with fuzzy techniques in content determination can actually show its highest usage potential.

Resuming our use case, the content determination task has now been succesfully developed by the fuzzy and NLG experts. As a result, the current system is able to extract content messages which contain both the general and extended information for the target reports (Fig. \ref{example}). Furthermore, the NLG experts have implemented the document structuring stage, and the system is also able to organize the content messages into different high level structures (such as paragraphs) according to the general structure described in the report sample (Fig. \ref{example}).

The current challenge for the NLG system developers is to decide how to lexicalize the content messages, i.e., which words and expressions should be used to communicate the information contained in the content determination messages. For some content parts this should be rather direct, such as the rainy days count message (Fig. \ref{rain_count_message}), but the introduction of the fuzzy content messages expands the possibilities and choices which can be made in this regard.

For instance, retaking the general temperature message from the content determination example in Fig. \ref{content_determ_quant}, which essentially consists in a type-I quantified sentence, the system developers may choose several alternatives. The first and most direct alternative is, if we obviate the domain language requirements, to express it as ``nearly all the days of the period were cold''. In this sense, a fuzzy quantified statement which follows a standard protoform can easily be matched with a natural language expression whose subject is ``Q of Xs'' and predicate is ``are A''. Thus, fuzzy quantified statements can also be considered as ``proto-lexicalization'' messages, since they are built using linguistic concepts which can already be directly expressed as natural language sentences.

However, as it was shown in Fig. \ref{protovstext} in Section \ref{sec:lddandnlg}, in real applications the expressions must be adapted to the domain language requirements, and converting a protoform-like statement into an actual appropriate natural language expression is not trivial in many cases. Thus, the NLG system developers of our use case decide to lexicalize the general temperature message as ``the period was predominantly cold''\footnote{Note that we are using actual natural language expressions to illustrate lexicalization, but this task actually involves creating new sets of structures which define an intermediate syntax between the original content messages and the actual text \cite{nlg_building_nlg}.}. Moreover, the lexicalization task is implemented so that every time a ``nearly all'' quantifier appears in the temperature message, the expression ``was predominantly'' is used systematically, as long as the fulfillment degree of the message is high (which is usually determined by establishing a certain threshold).

In fact, without taking into account the fulfillment degree, the lexicalization task in our use case would not be too different from standard lexicalization approaches. The semantics of the general content messages for temperature and rain is a combination of the linguistic terms which characterize the fuzzy quantified statements and their associated fulfillment degrees, and everything must be taken into account when performing lexicalization. Another simpler, but perhaps more illustrative example of this is the lexicalization process for exceptional temperature periods (which deviate from the general trend) in the extended information part of the report in Fig. \ref{example}.

\begin{figure}[h]
\centering
\includegraphics[width=0.7\columnwidth]{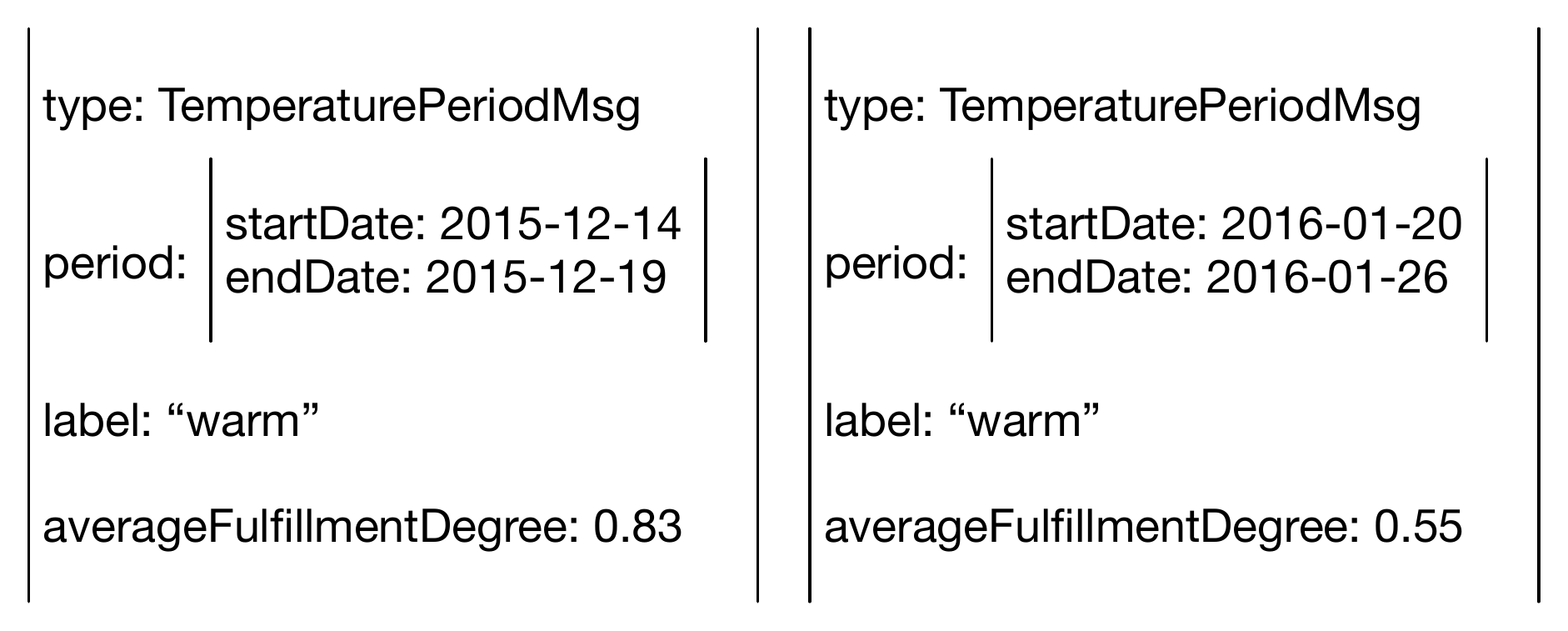}
\caption{Content determination messages for the temperature periods element. Although they share the same linguistic label, the semantics of the messages are influenced by their fulfillment degree.}
\label{lexicali_temp}
\end{figure}

Let us suppose that the NLG system development team has implemented as part of the content determination task a fuzzy algorithm which extracts relevant temperature periods from the input data. Each period is characterized by its start and end dates, a temperature label and the average fulfillment degree calculated by evaluating the label against the input data. Suppose now that for a given input data set, the system obtains two messages for a same temperature label, as Fig. \ref{lexicali_temp} shows.

The most obvious solution in this case would be to lexicalize both messages as ``warm periods''. However, this way we would be omitting the semantic contribution of the fulfillment degree and nullifying the flexibility that fuzzy sets provide. Once more, to grasp the contribution of a fulfillment degree to the semantics of a linguistic term such as ``warm'' is a problem the NLG system experts have to solve. For instance, ``warm'' periods with an average fulfillment degree in the range [0.5,0.75] could be lexicalized as ``warmish'' (not entirely warm), while periods in the range (0.75,1] are lexicalized as ``warm''. Content messages with an average fulfillment degree lower than 0.5 would already be discarded as part of the content determination task algorithm.

It is also possible that additional criteria should be taken into account in the case of Fig. \ref{lexicali_temp}, such as the fulfillment degree of contiguous labels for each message. Focusing on the ``warm $FD=0.55$'' message, one could consider computing the fulfillment degrees for contiguous ``cold'' and ``hot'' fuzzy labels for the same temporal period. This increases the possibilities, as a ``cold $FD=0.35$'' or a ``hot $FD=0.4$'' could be used in conjuction with ``warm $FD=0.55$'' to produce expressions such as ``warm/coldish'' or ``warm/hottish''.

The previous examples show how individual messages can be lexicalized in the context of using fuzzy approaches and illustrate some of the possibilities these techniques open in this regard. However, to simply convert single messages
into natural language expressions is not enough in most cases to produce actual texts. For instance, if the NLG system experts fed the linguistic realizer with the current lexicalization structures, the system would generate something similar to \textit{``The period was cold in general. The period was wet in general. There were 77 days with rain. There was a warm period from the 14th to the 19th of December. There was a warmish period from the 20th to the 26th of January...''}, which is perfectly correct in terms of ortography, syntax and grammar, but is repetitive and still lacks readability. The aggregation task plays an important role in addressing this issue. 

\subsubsection{Aggregation} \label{sec:agg}
In general, the aggregation task in NLG systems involves the creation of more complex sentences by merging simple phrase specifications \cite{nlg_building_nlg}. This task is usually depicted from a structural perspective (see Fig. \ref{pipeline}), and is usually addressed by using mechanisms such as simple conjunction, conjunction via shared participants, conjunction via shared structures, etc. For instance, the NLG experts in our use case may add several aggregation rules in order to merge some of the sentences which have been lexicalized, e.g. the general temperature and rain descriptions. Thus, ``the period was predominantly cold'' and ``the period was predominantly wet'' could be merged via shared participants (they share the same subject, ``the period'') as ``the period was predominantly cold and predominantly wet''. Alternatively, the previous messages could also be merged through a shared structure mechanism as ``the period was predominantly cold and wet''.

\begin{figure}[h]
\centering
\includegraphics[width=\columnwidth]{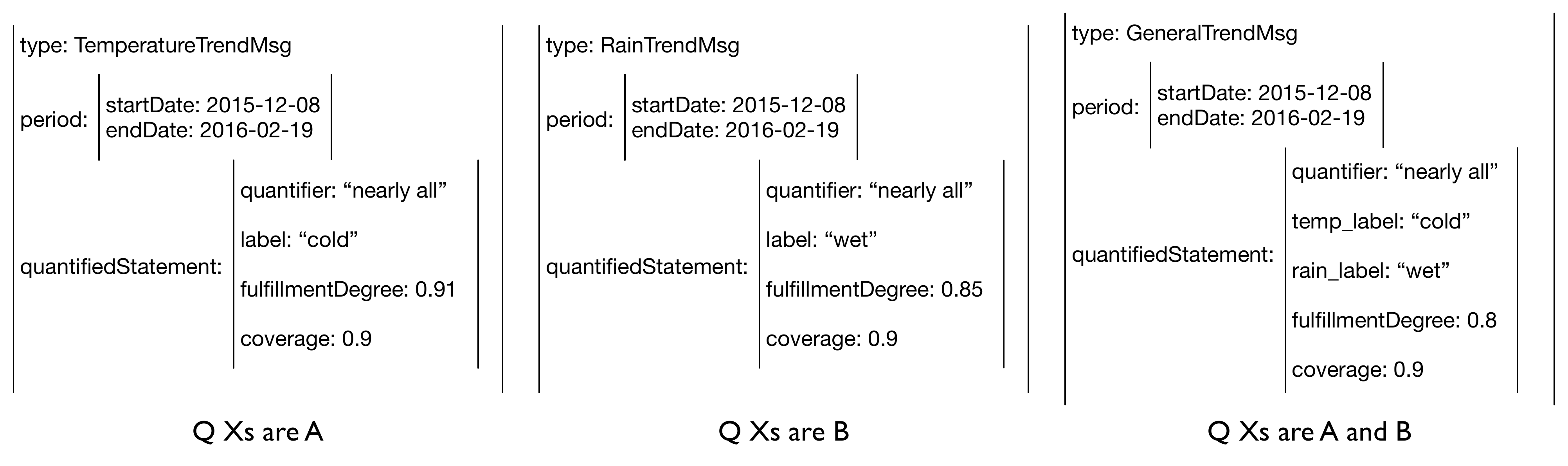}
\caption{Content determination messages for single temperature and rain content elements, and possible definition of a message which represents the logical aggregation of both properties by means of a fuzzy t-norm.}
\label{aggregation1}
\end{figure}

While these mechanisms operate at an strictly syntactical or structural level, they resemble similar content aggregation mechanisms which are present in fuzzy sets theory and LDD. These are usually modeled by means of t-norm operators, such as the minimum or the product. For example, if we backtrack the generation of the ``the period was predominantly cold and wet'' sentence, two alternatives now emerge. The first one, which has already been described, is a conjuctive aggregation operation via shared structures. The second one, however, could have been performed during the content determination task by obtaining a single quantified sentence which aggregates both temperature and rain properties (``Q Xs are A \textbf{and} B''), such as ``near all the period days were cold and wet'. However, ``and'' in this case would not be a simple conjunction which merges two sentences, but rather a t-norm operator.

This arises some questions. Consider the case where the sentence ``the period was predominantly cold and wet'' is generated using the two distinct aforementioned approaches (see Fig. \ref{aggregation1}). To which extent their meaning would be equivalent? And to which extent does using a specific fuzzy aggregation operator influence the semantics of the obtained stamement and its equivalence with the other sentence? Again, to decide which specific operator to use will be a non-trivial decision the experts will have to make if they choose to aggregate properties using fuzzy operators, especially when other kind of non-standard operators have been shown to better resemble the way humans aggregate information \cite{bib_zimmermann,bib_yager_owa}. Thus, using such fuzzy aggregation techniques at a content determination level may imply additional issues, but this can also help alleviate the complexity of the subsequent lexicalization and aggregation subtasks.

\begin{figure}[h]
\centering
\includegraphics[width=0.9\columnwidth]{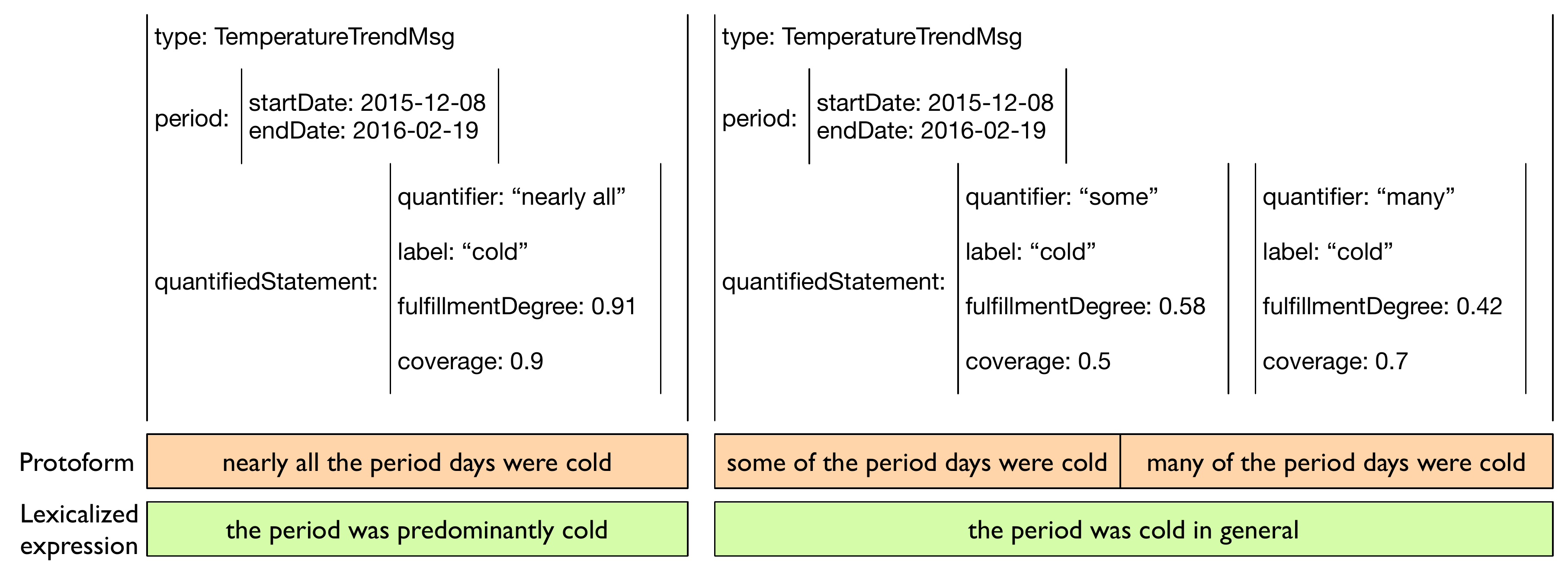}
\caption{Content determination messages for the general temperature description. The one on the right depicts a situation where more than one quantified statement is considered.}
\label{aggregation2}
\end{figure}

Another interesting case of aggregation is also introduced by the use of LDD techniques within the NLG pipeline. Let us consider a situation in our use case where the general temperature trend message is composed of two submessages (which means that the fulfillment degree is spread among more than one quantifier), as Fig. \ref{aggregation2} shows. The second message in Fig. \ref{aggregation2} is composed of two different statements, namely ``some of the period days were cold'' and ``many of the period days were cold''. Although they are not incompatible from a fuzzy logic perspective (both fulfillment degrees are close to 0.5), to lexicalize both independently would probably lead the reader to an important confusion state. Likewise, if we aggregate them using the aforementioned conjunctive linguistic mechanisms, a sentence like ``some and many of the period days were cold'' would be generated, which does not improve the situation much.

In this problem the key lies in providing an expression which reflects the semantics of the message in a proper way. Intuitively, the whole message could be expressed as ``between some and many of the period days were cold'' or ``some or many period days were cold'', among others, which are far from the domain language requirements. Thus, in order to cut corners, the experts in our use case decide to lexicalize it as ``the period was cold in general''. Other semantic interpretations are possible, since it can also be considered that the quantifier ``some''' is more inespecific than ``many'' and, in terms of their fuzzy sets interpretation it should hold that $some \supset many$. Under these considerations consistent lexicalizations can be either to state the more inespecific quantifier $some=some \cup many$ if the aim is to cover as much data as possible or $many=some \cap many$ if providing the most specific information is preferred to data covering. One could say that this latter specific problem is more related to lexicalization (choosing a natural language expression for a given more complex message) than aggregation, but even the well-defined NLG architecture by Reiter and Dale admits a high flexibility of opinions in similar task placement issues \cite{nlg_building_nlg}.  

\subsubsection{Referring Expression Generation} \label{sec:reg}
The task of generating referring expressions emerges as a beast of its own and may well be the less understood and most actively researched subtask within NLG. Although it is directly to related to lexicalization, it focuses on a more specific problem. In short, its purpose is to identify entities (thing, being, event...) within a discourse and generate expressions which provide such identification (referring expressions). These are usually expressed in the form of noun phrases, or surrogates for a noun phrase.

\begin{figure}[h]
\centering
\includegraphics[width=0.7\columnwidth]{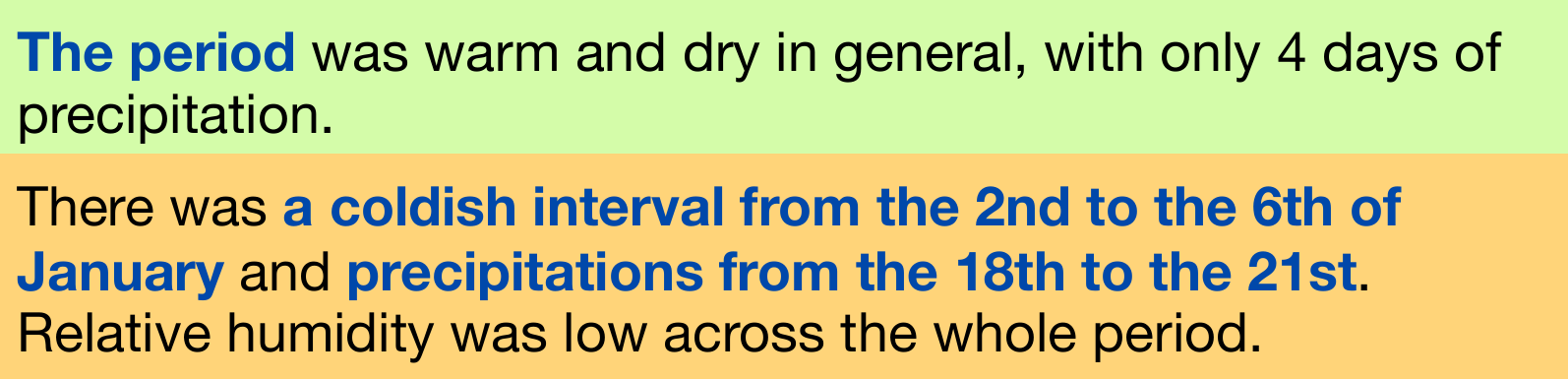}
\caption{Report example from Fig \ref{example}, including highlighted referring expressions.}
\label{reg_example}
\end{figure}

If we follow the previous definition, some referring expressions can be identified in the report example from Fig. \ref{example}, such as ``the period'', or ``a coldish interval from the 2nd to the 6th of January'' (Fig. \ref{reg_example}). The case of ``the period'' is perhaps one of the simplest referring expression examples which can be found, as the reader of the report will certainly know which time interval the report is referring to (there is only one period and thus no features are needed to further identify such entity). However, in ``a coldish interval from the 2nd to the 6th of January'' we find a more interesting referring expression, which identifies an specific event by means of a fuzzy property (cold, expressed as ``coldish'') and a temporal expression (which in this case is crisply defined, but could also be a fuzzy temporal expression in other situations \cite{bib_gatt_portet_uncertain}, e.g. ``towards the beginning of January'').

\begin{figure}[h]
\centering
\includegraphics[width=0.8\columnwidth]{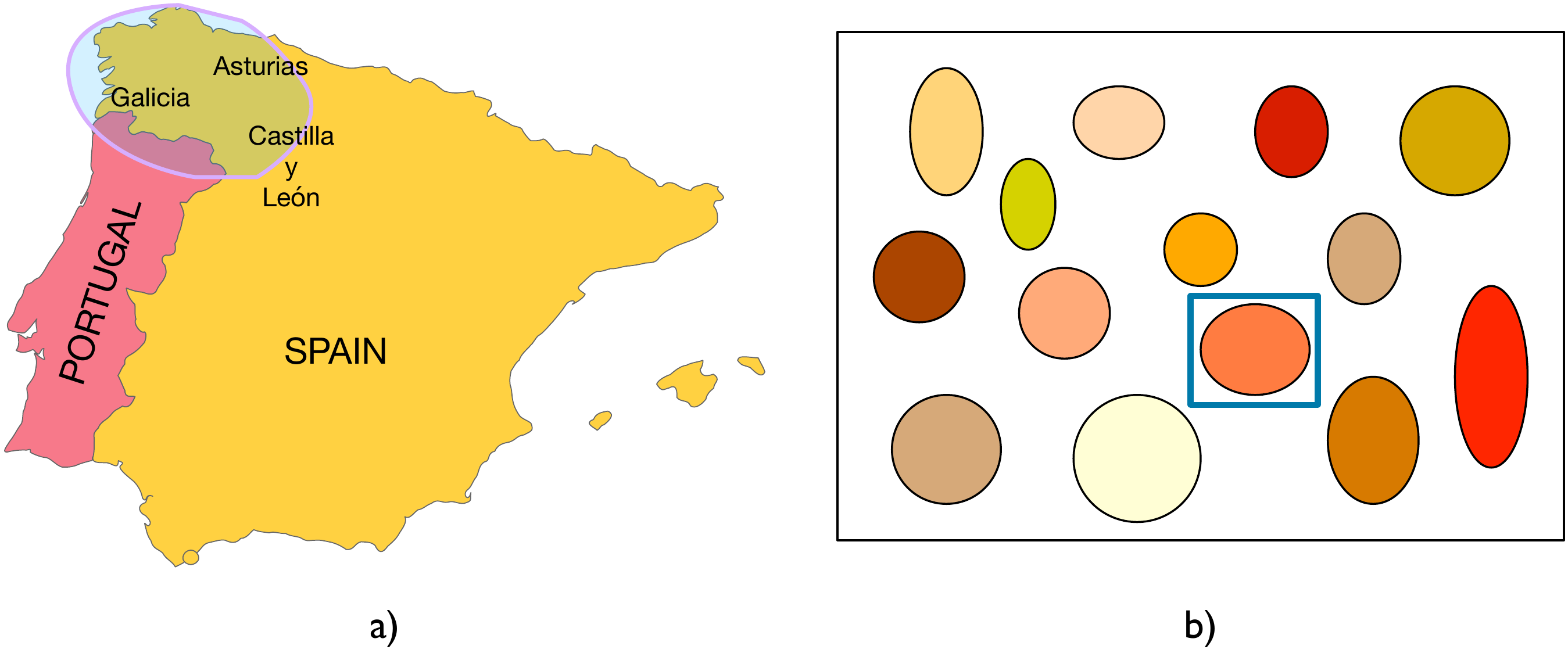}
\caption{\textbf{a)} Example of a geographical region in the problem of generating geographical referring expressions. \textbf{b)} Example of referring expression generation problem in an image. Which set of fuzzy properties would allow to generate an expression which distinguishes the circle within the blue square in an as unambiguously as possible manner (needless to say, supposing the blue square does not exist)?}
\label{gre}
\end{figure}

Fuzzy sets and LDD allow for the inclusion of fuzzy properties which can be used for the task of referring expression generation. In this regard, another interesting open research problem in NLG is the generation of geographical referring expressions which help identify within a single discourse relevant events through the use of geographical descriptors. For instance, consider the highlighted region in Fig. \ref{gre} a). Could this region be described as ``NW Spain'', ``Galicia, Asturias and part of Castilla y León'', ``to the north of Portugal'' or simply ``near Galicia''?

Such geographical referring expressions have been generated through crisp approaches classically, such as the ones produced by the RoadSafe system \cite{Turner2008, nlg_roadsafe2}, which is able to generate complex expressions from geographical data like ``in some far southern and southwestern places'' \cite{bib_ross_3}. However, the use of crisp boundaries between the descriptors leads to situations where even if one data point is only slightly below another point, one could be computed as ``north'' and the other one as ``south''. Fuzzy sets and LDD can help address such problems by modeling fuzzy geographical descriptors which can be combined to generate similar geographical referring expressions which consider the graduality of such geographical concepts \cite{bib_glockner}.

To characterize discourse entities using fuzzy properties for the task of generating referring expressions will also have a deep impact in how referring expression algorithms should be approached (in a similar manner to how the other tasks within the NLG pipeline are affected by this, since increasing flexibility by using gradual properties also increases the complexity which must be managed). It must also be noted that this complexity will be given by the kind of domain entities which must be referred to. For instance, in D2T systems which generate texts from time series data, language tends to be simpler \cite{nlg_datatotext} and relevant information such as events can be clearly identified through the use of temporal expressions. In others, such as the aforementioned geographical NLG systems or visual systems in general \cite{bib_rita_images}, this complexity will be higher (see Fig. \ref{gre} b)).

\section{Discussion and Conclusions} \label{sec:conclusions}
\subsection{Additional remarks}
We have studied how fuzzy sets and LDD can play an important role in bringing imprecision and uncertainty management to NLG systems. While they seem to be directly usable for content-related tasks, we have also explored how a structure-related task, such as aggregation, could also be approached in some cases through fuzzy means.

In this regard, we would like to retake the idea proposed by Kacprzyk and Zadro\.zny in \cite{kacprzyk_n2} about the creation of new kinds of protoforms \cite{wilbik_new_protoform}, which was discussed in Sec. \ref{sec:lddandnlg}. Although we did not describe in this paper document structuring as an NLG task which could be directly approached by LDD or fuzzy means, we believe that this could serve as inspiration to design richer protoforms.

While document structuring deals with the organization of content messages into higher level structures such as paragraphs, this sorting operation also takes into account other kind of relationships among these messages, namely discourse relations, such as contrast, elaboration or narrative sequences. It is not difficult to imagine extended protoforms which model somehow these discourse relations. Consider, for instance, the following protoforms
\begin{equation}
``Q \  Xs \  \text{are} \  A\text{, but} \  R \  Ys \  \text{are} \  B"
\label{contrast_eq}
\end{equation}
\begin{equation}
``Q \  Xs \  \text{are} \  A\text{, especially} \  R \  Ys \  \text{are} \  B"
\label{empha_eq}
\end{equation}
which model a contrast (\ref{contrast_eq}) and an emphasizing (\ref{empha_eq}) relation between two basic protoforms. To obtain protoform-like statements such as ``Most of the days in 2015 were warm, but most of the days in November 2015 were very cold'' or ``Most places in Southern France suffered from strong precipitations, especially most of the SW coast suffered from intense floods''. Such kind of protoforms would also have an impact in how the text should be structured, as a relation holds between their more basic elements.

Although this kind of contrast and emphasizing relationships between different content elements were considered in \cite{bib_cisti13} from a fuzzy perspective, further research could be made to adapt this kind of discourse relations into new types of protoforms. This would also mean that the semantics of such relations could be approached in a fuzzy way (e.g. a fuzzy measure of contrast among different labels of a same linguistic variable, based, for instance, on a standard measure of antonimy).

\subsection{Evaluation of NLG systems using fuzzy approaches} \label{sec:eval}

The evaluation of NLG systems which include uncertainty management through fuzzy sets will be another interesting research scope, as the flexibility such techniques provide implies that additional decisions will have to be taken (e.g. as in the examples shown in Sections \ref{sec:cd}, \ref{sec:lex} and \ref{sec:agg}). To assess all these design decisions during the conception and development of an NLG system will probably be unfeasible, which opens the possibility of including such assessment during the subsequent standard evaluation process of the system \cite{nlg_gattevaluation}.

In this regard, we would like to differentiate the concept of ``evaluation'' used in a LDD perspective from the one used in an NLG context. Evaluation of LDD approaches has traditionally consisted in the usage of different criteria, such as the fulfillmente degree or coverage used in the use case in Sec. \ref{sec:across}, which allow to rank and select the most appropriate candidate statements generated by an LDD algorithm. Thus, such evaluation would be performed as part of the content determination process.

To evaluate an NLG system is a task which is usually done once the system is fully functional and is able to generate actual texts. In this regard, two different types of evaluations may be performed: \textit{i)} intrinsic evaluations, which focus on the quality and appropriateness of the texts generated by the system, and \textit{ii)} extrinsic evaluations, which try to measure the extent to which the system is actually useful and impactful for its users \cite{bib_role_ldd_nlg}.

In our opinion, many of the decisions taken as part of the NLG system development process due to the use of fuzzy approaches could be assessed inherently through intrisinc evaluations (which usually consist in obtaining feedback from the domain experts in the form of questionnaires or text-free comments). In some cases, it could also be interesting to create different versions of a same NLG system and assess their differences (for instance, to study the problem of aggregation at a content determination level using fuzzy operators and the aggregation at a structural level).

\subsection{Terminology}
Another issue, which is indirectly related to the research questions explored here but may very well become a problem in the future, is the terminology used in the literature to refer to each discipline involved in the tasks here described. We have tried to keep here a clear distinction between LDD, NLG and D2T and, although the two latter terms are well established in terms of usage, LDD has been named differently or used with distinct meanings in the literature. LDD was originally conceived as ``linguistic summarization of data'' \cite{bib_Yager} and this name is still used in many fuzzy sets research papers. While we believe that it represents well the purpose of the field which represents, this terminology may confuse readers from other disciplines, as summarization is a well-known discipline in NLG and NLP (generating texts which summarize larger documents), which is totally unrelated to LDD.

Other authors are considering LDD as an alternate approach which actually reunites NLG/D2T and what we understand as LDD \cite{bib_ldd_time_series}. This is an interesting proposal which fits well the idea of reuniting both paradigms, but a consensus should be achieved to avoid further confusion. Perhaps, the most surprising fact in this terminology discussion is that none of the names used until now (``linguistic summarization of data'', ``linguistic description of data'') explicitly emphasizes the fuzzy nature of the techniques and operations that this discipline encompasses.

\subsection{Conclusions}
The use of fuzzy sets and linguistic descriptions of data to provide imprecision and uncertainty management capabilities in NLG systems is a promising research line which, as we have discussed in this paper, has many ramifications. Although this kind of techniques seem to fit primarily in content-related tasks, the diversity of problems involved in such tasks allows for many possibilities. Furthermore, even more structure-focused NLG tasks such as aggregation or document structuring could also benefit from fuzzy sets and LDD. Table \ref{summary} summarizes the most relevant potential applications of these techniques in NLG.

\begin{table}[h]
\centering
\caption{Summary table of potential applications of fuzzy sets and LDD in NLG}
\begin{tabular}{| p{0.25\textwidth} | p{0.65\textwidth}|}
\hline
\textbf{NLG Task} & \textbf{Application of fuzzy sets and LDD} \\ \hline
\noalign{\smallskip} \hline
Content determination & Use of fuzzy quantified statements and fuzzy properties to obtain imprecise information, e.g. ``Most of the period days were dry'' or ``Temperatures were high during the first fortnight''.\\ \hline
Document structuring & Modeling discourse relations through complex protoforms: for instance, contrast or emphasizing relationships, e.g. ``The month was predominantly warm but there was a cold period towards the end''. \\ \hline
Lexicalization & Based on the obtained fuzzy information, decide how to express it in natural language (Fig. \ref{protovstext}). How does a fulfillment degree influence the semantics of a given expression? \\ \hline
Aggregation & Could be performed in some cases by using standard aggregation operators at a content determination level, e.g. ``The month was cold and dry''. However research should be made to determine the equivalence of structural aggregation and the use of fuzzy operators. \\ \hline
Referring expression generation & Define fuzzy properties which allow to identify certain entities in the discourse, fuzzy referring expression generation algorithms. \\ \hline
Linguistic realization & No apparent applications. \\ \hline
Structure realization & No apparent applications. \\ \hline 
\end{tabular}
\label{summary}
\end{table}

Likewise, the research on NLG systems will also help bring fuzzy sets and LDD closer to real applications where uncertainty plays a key role. In this regard, as discussed in Section \ref{sec:lddandnlg}, one of the main challenges will be to embrace and adapt the empirical approaches usually performed during the conception of NLG systems.

The final purpose of this strong collaboration between these both major fields in the artificial intelligence community is to provide better systems which, in the context of Data Science, produce more human-friendly information in the form of natural language texts while managing the vagueness and imprecision included in the semantics underlying such information. Within this context, D2T/NLG systems, either alone or as a complementary support to visualization, will allow to improve the understanding of large data sets in many application domains and bring data closer to people. Empiric studies \cite{nlg_metofficedatatotext} show that visual information alone is not always capable of adequately communicating relevant information about data to users. In this regard NLG and D2T are descriptive approaches that, combined with sound analysis techniques, are starting to prove to be valuable complementary informative tools in the Data Science realm.

As future work, the authors will follow the tips here provided and research the use of fuzzy sets in the research and development of NLG systems. Particularly, we will focus on \textit{i)} researching the generation of geographical referring expressions by means of fuzzy sets, \textit{ii)} exploring how to better integrate LDD within D2T in terms of content determination and lexicalization tasks and \textit{iii)} identifying new potential application domains where the use of LDD+NLG may prove useful.

\section*{Acknowledgments}
This work was supported by the Spanish Ministry for Economy and Competitiveness (grant TIN2014-56633-C3-1-R) and by the European Regional Development Fund (ERDF/FEDER) and the Galician Ministry of Education (grants GRC2014/030 and CN2012/151). Alejandro Ramos Soto (A. Ramos-Soto) is supported by the Spanish Ministry for Economy and Competitiveness (FPI Fellowship Program) under grant BES-2012-051878.

\bibliographystyle{IEEEtran}
\bibliography{IEEEabrv,bibliography}

\end{document}